\documentclass[]{innovator}
\usepackage{enumitem}
\usepackage[toc,page,header]{appendix}
\usepackage[utf8]{inputenc}
\usepackage[T1]{fontenc}
\usepackage{hyperref}
\usepackage{url}
\usepackage{graphicx}
\usepackage{booktabs}
\usepackage{amsfonts}
\usepackage{nicefrac}
\usepackage{makecell}
\usepackage{microtype}
\usepackage{amsmath}
\usepackage{etoolbox}
\usepackage{lipsum}  
\usepackage{minitoc}
\usepackage{tablefootnote}
\usepackage{threeparttable}
\usepackage{wrapfig}
\usepackage{appendix}
\usepackage{multirow}
\usepackage{ulem}
\useunder{\uline}{\ul}{}
\usepackage{colortbl}
\usepackage{longtable}
\usepackage{color}

\title{Innovator: Scientific Continued Pretraining with Fine-grained MoE Upcycling} 

\author[1,2]{Ning Liao}
\author[1,2]{Xiaoxing Wang}
\author[2,3]{Zehao Lin}
\author[1]{Weiyang Guo}
\author[1]{Feng Hong}
\author[1,2]{Shixiang Song}
\author[1]{Geng Yu}
\author[1]{Zihua Zhao}
\author[1]{Sitao Xie}
\author[1]{Longxuan Wei}
\author[1]{Xiangqi Jin}
\author[1]{Xiaohan Qin}
\author[1]{Jiale Ma}
\author[2]{Kai Chen}
\author[1]{Jiangchao Yao}
\author[1]{Zhouhan Lin}
\author[1]{Junchi Yan}
\author[2,3]{Zhiyu Li}
\author[2,3]{Feiyu Xiong}
\author[1]{Yanfeng Wang}
\author[1]{Linfeng Zhang}

\affiliation[1]{Shanghai Jiao Tong University}
\affiliation[2]{MemTensor (Shanghai) Technology Co., Ltd.}
\affiliation[3]{Institute for Advanced Algorithms Research, Shanghai.}

\abstract{
A large language model (LLM) with knowledge in both scientific and general tasks is the foundation of science general intelligence. However, directly continued pretraining an LLM using science data usually leads to catastrophic forgetting, which indicates severe degradation in general ability. In this report, we present \textbf{Innovator}, which solves this problem by upcycling a pre-trained dense LLM into a fine-grained Mixtures-of-Experts model during continued pretraining, where different experts are expected to learn science knowledge in different disciplines, and a shared expert is utilized for general tasks. Innovator introduces a four-stage upcycle training paradigm: (1) Scientific Expert Induction on discipline-specific data, (2) Fine-grained Expert Splitting via FFN dimension decomposition, (3) Science-Aware Routing warmup, and (4) Generalist-Scientist Integration training on hybrid datasets. Such a paradigm enables knowledge in the general domain, and different scientific disciplines can be decoupled, avoiding the negative influence among knowledge in different domains. 
With 53.3B total parameters and 13.3B activated, Innovator extends Qwen2.5-7B using a shared general expert and 64 specialized scientific experts with 8 activated. Trained on 300B tokens with tri-level quality-controlled data, Innovator achieves 25\% average improvement across 30 scientific tasks with a win rate as 70\%, while retaining 99\% performance in general tasks. Furthermore, \textbf{Innovator-Reason}, which is post-trained from Innovator for reasoning boosting, exhibits excellent reasoning performance in solving complex scientific problems with improvements over 30\%.
}

\begin{document}
\maketitle

\section{Introduction} 
The relentless human pursuit of science has continually driven the transformations and progress within human society. In the last several years, Large Language Models (LLMs) have evolved beyond  \textbf{chatbots} into \textbf{reasoners}, and then obtained the ability to use tools as \textbf{agents}. Consequently, enabling LLMs to work as scientific \textbf{innovators} has become the most paramount challenge. The first step for this target begins with a model that incorporates substantial knowledge across a wide range of disciplines of science, encompassing complex scientific data such as molecular formulas, reaction equations, SMILES, and charts, while still preserving the original ability of general tasks.

Some previous works have tried to tackle this challenge by continued pretraining or domain-specific supervised finetuning~\cite{zhang2024comprehensive}. They mainly focus on exploring the potential of LLMs to assist in understanding complex scientific literature, extracting knowledge, and solving domain-specific problems across disciplines. For example, Galactica~\cite{taylor2022galactica} is trained on a broad scientific corpus to enhance reasoning over scientific literature, formulas, and structured knowledge, while GeoGalactica~\cite{lin2023geogalactica} adapted this framework to geosciences through domain-specific pretraining. SciDFM~\cite{sun2024scidfm} has learned from numerous  scientific papers and books with domain-specific databases. It is capable of conducting college-level scientific reasoning, understanding molecules and amino acid sequences. In the field of chemistry, ChemLLM~\cite{di2024chemllm} and ChemCrow~\cite{m2024augmenting} integrate LLMs with expert tools to assist in synthesis planning and molecule discovery. Despite these promising efforts, these works are often tailored to individual domains, lacking generalizability across diverse scientific disciplines. Besides, training LLMs on scientific data usually severely harms their performance in general domains, since the scientific notations and the related expressions exhibit a significant difference compared with the general data, making the LLM become a domain-specific expert, while losing the ability as a cross-domain scientist. As a result, injecting scientific knowledge into LLMs without a negative influence on their general ability has become the most challenging problem.

Recently, sparse Mixture-of-Experts~\cite{shazeer2017outrageously} (MoE) models,
such as Mixtral~\cite{jiang2024mixtral}, Deepseek-v3~\cite{liu2024deepseek}, and Qwen-series MoE~\cite{team2024qwen2,yang2025qwen3}, have demonstrated significant advancements, especially in their ability to decouple knowledge in different domains by using different experts.
These models integrate MoE layers, which consist of several Feed-Forward Network (FFN) experts and a trainable router, as a replacement of traditional single FFN layers. By selectively activating only a subset of experts for each input, MoE models achieve high performance with improved computational efficiency. However, training MoE models from scratch requires enormous computational resources~\cite{riquelme2021scaling, fedus2022switch, gale2023megablocks} and often suffers from instability during training~\cite{zoph2022st, puigcerver2023sparse}. To overcome these challenges, a promising solution has been proposed in the form of upcycling pre-trained dense models into MoE models~\cite{komatsuzaki2022sparse}. This approach initializes the experts within the MoE layers using the weights from the pre-trained FFN. By leveraging the existing knowledge embedded in the dense model, upcycled MoE models facilitate more efficient training and deliver competitive results even with limited computational budgets~\cite{wei2024skywork,he2024upcycling,lin2024moe}. Given its straightforwardness and efficacy, upcycling has been widely applied across various fields~\cite{komatsuzaki2022sparse,huang2025ders,he2024upcycling,lin2024moe, jiang2024medmoemixturedomainspecificexperts}. 

In this report, we propose \textbf{Innovator}, which has been continuously pretrained on hybrid general and scientific data by upcycling from a dense model into a fine-grained MoE model, where the knowledge in different disciplines is expected to be learned by different experts. Instead of directly copying multiple FFNs in the dense model as different experts, Innovator introduces a four-stage paradigm. Firstly, it induces multiple coarse-grained experts by training on data from each single scientific discipline, which are then utilized as the initialization in the following stage. Then, it splits the coarse-grained experts into multiple fine-grained experts by directly splitting along the higher dimension of FFNs, which enables the ability of different scientific domains to be further decoupled. Thirdly, with an original dense model as a shared expert and multiple fine-grained experts as expert candidates, Innovator is further trained on its router to accurately dispatch data into corresponding scientific experts. Finally, the entire model is further trained on a mixture of general and scientific data for generalist and scientist integration.

Our experiments demonstrate that Innovator shows significant improvements in scientific understanding and reasoning, with almost no degradation in its general ability. Concretely, after training on 300B tokens, it improves model performance on 30 scientific tasks by around 25\% while preserving 99\% performance in general tasks. Moreover, through post-training on Innovator, the resulted model, namely Innovator-Reason, achieves a great performance advantage in both non-scientific and scientific reasoning evaluation with logical and concise thinking process.

\section{Architecture} 
Innovator is an MoE model extended from the Qwen2.5-7B~\cite{Qwen2.5} dense model. It contains 53.3B total parameters with 13.3B activated for each token. Similar to the Qwen2.5-7B, the attention layers adopt Grouped Query Attention~\cite{GQA}, SwiGLU~\cite{swiglu}, Rotary Positional Embedding~\cite{rope}, and RMSNorm~\cite{rmsnorm} with pre-normalization. 
The MoE layer consists of 1 shared general expert and 64 scientific experts with 8 experts activated each time. The architecture of the shared general expert is the same as the Feed-Forward Network (FFN) of the Qwen2.5-7B dense model. The scientific experts are implemented by fine-grained expert segmentation, and its intermediate size is one eighth the one of the shared expert. In addition, we also design a science-aware routing strategy to dispatch data from different subjects to the corresponding scientific experts.
For tokenization, we utilize Qwen2 tokenizer~\cite{team2024qwen2} adopted in the Qwen2.5-7B with a vocabulary size as 152064.

\section{Data Preparation} 
\begin{figure}[t!]
\centering
\includegraphics[width=1\textwidth]{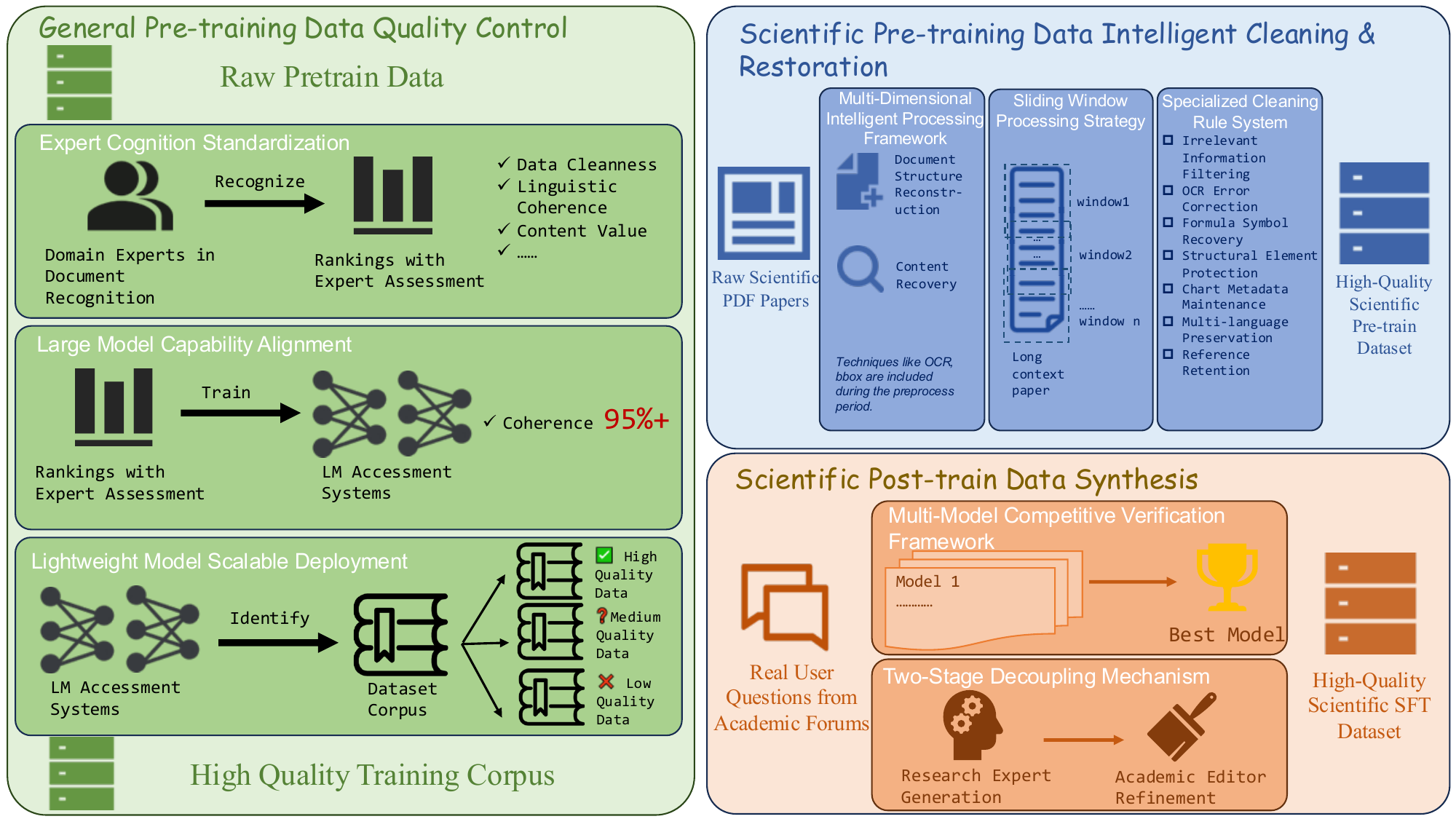}
\caption{The overview of the proposed framework for data preprocessing. It covers distinct procedures tailored for general pre-training, scientific pre-training, and post-training phases.}
\label{fig:data_engineer}
\end{figure}

During scientific continued pretraining of LLMs, data quality critically determines the upper bounds of both knowledge retention and acquisition \cite{gururangan2023scaling, xie2023data}. Traditional methods exhibit significant limitations when processing petabyte-scale scientific corpora: heuristic rules (e.g. C4 dataset~\cite{dodge-etal-2021-documenting}, Wanjuan dataset~\cite{qiu2024wanjuan}) are efficient but lack domain specificity, while single classifiers (e.g. RedPajama v2~\cite{weber2024redpajama}, RefinedWeb~\cite{penedo2023refinedweb}, FineWeb-Edu~\cite{penedo2024fineweb}) cannot capture the multidimensional nature of scientific data quality. 
To this end, we introduce a comprehensive tri-level aligned framework encompassing ``human-expert-defined, strong-model-annotated, and small-model-filtered'' modules specifically designed for scientific continued pretraining as shown in Fig.~\ref{fig:data_engineer}. The framework builds upon foundational techniques including heuristic filtering and full-scale MinHash deduplication, enabling intelligent multidimensional quality assessment across general and scientific domains.

Specifically, our approach features a hierarchical knowledge-transfer architecture that achieves closed-loop quality control from expert cognition to large-scale deployment. To support the four-stage upcycle training paradigm of Innovator and post-training, we meticulously curate datasets spanning general pretraining, scientific pretraining, and post-training phases. The detailed data preparation pipeline is described below.

\subsection{General Pre-training Data: Multi-Level Alignment Quality Control}
To preserve general capabilities during scientific continued pretraining, we develop a three-level alignment system that extends beyond conventional filtering and deduplication approaches \cite{lee2022deduplication}:

\begin{itemize}
    \item \textbf{Level 1: Expert Standardization}: 
    Domain experts establish multi-dimensional quality standards encompassing cleanliness, coherence, and content value through systematic sample annotation. This rigorous standardization process defines foundational quality metrics that serve as the bedrock for subsequent automated processing stages in our continued pretraining pipeline.
    
    \item \textbf{Level 2: Model Alignment}: 
    Advanced LLMs are aligned with expert standards through targeted prompt engineering and instruction tuning, achieving >95\% labeling consistency with human annotations. This alignment stage bridges expert human judgment with scalable automated processing, enabling high-fidelity quality assessment for large-scale continued pretraining data.
    
    \item \textbf{Level 3: Lightweight Deployment}: 
    Knowledge distillation produces efficient bilingual quality assessment models (85.99\% CN/86.50\% EN accuracy) for industrial-scale filtering. These compact models enable real-time, cost-effective quality control across petabyte-scale datasets during the continued pretraining process.
\end{itemize}

Our quality control system processes extensive raw corpora into high-quality tokens for continued pretraining.

\subsection{Scientific Pre-training Data: Intelligent Cleaning and Restoration System}
For scientific continued pretraining, we develop an LLM-based processing system specifically designed for high-quality reconstruction and curation of scientific literature:

\begin{itemize}
    \item \textbf{Multi-Dimensional Processing}: 
   We implement an integrated pipeline for document reconstruction encompassing page-order correction via bounding box analysis, LaTeX formula reconstruction, and page-break repair. This comprehensive approach addresses the complex structural and content integrity challenges inherent in scientific documents, ensuring maximum information fidelity for continued pretraining on scientific knowledge.
    
    \item \textbf{Sliding Window Strategy}: 
     We employ context-preserving segmentation with optimized overlaps specifically designed for cross-page formulas and tables. This strategy maintains semantic continuity of fragmented scientific elements, preventing information loss that commonly occurs with conventional segmentation approaches during scientific continued pretraining.
    
    \item \textbf{Scientific Cleaning Rules}: 
     Our specialized rule set encompasses irrelevant information filtering, OCR error correction, formula symbol recovery, structural element protection, chart metadata maintenance, multi-language preservation, and reference retention. These domain-specific rules are designed to preserve the accuracy and integrity of complex scientific notations and document structures essential for effective scientific continued pretraining.
\end{itemize}
Our processing system handles extensive scientific literature, producing billions of high-quality tokens with 68\% quality improvement (3.06→5.16 on our internal scientific literature quality assessment scale), providing a robust foundation for scientific domain continued pretraining.

\subsection{Scientific Fine-tuning Data: Two-Stage Generation for Instruction Synthesis}
To support supervised fine-tuning before post-training, we develop a "Generate-Refine" two-stage synthesis system for creating high-quality scientific data:

\begin{itemize}
\item \textbf{Two-Stage Decoupling}: 
We employ distinct roles where a "research expert" persona generates scientific content while an "academic editor" persona refines quality and presentation. This decoupled approach avoids creative-control interference, ensuring both comprehensive coverage of scientific knowledge and precision in academic presentation for effective post-continued-pretraining instruction tuning.

\item \textbf{Multi-Model Verification}: 
Multiple LLMs cross-validate generated outputs \cite{zheng2023judging}, producing diverse supervised fine-tuning datasets for comprehensive evaluation. This robust verification framework enhances the reliability of synthesized instruction data, providing high-quality resources for scientific capability refinement after continued pretraining.
\end{itemize}

Our synthesis system processes millions of scientific data points \cite{alpaca}, enabling effective scientific capability enhancement following the continued pretraining phase.

\subsection{Post-Training Data: Boosting Scientific Reasoning}
To further enhance scientific reasoning capabilities after continued pretraining, we construct a comprehensive post-training dataset spanning scientific tasks, mathematics, coding, and general capabilities. This dataset supports reinforcement learning to refine the model's performance on complex scientific problems following the four-stage upcycle training. Details of the dataset composition are presented below. 
\begin{table}[t!]
  \centering
  \small
  \caption{Example data overview for the four scientific RL datasets: Physics, Chemistry, Mol2Caption, and Caption2Mol. Each row lists an illustrative example, its ground-truth label, and the dataset size.}
  \label{tab:rl_datasets_transposed}
  \begin{tabular}{p{2.0cm}|p{3.0cm}|p{4.5cm}|p{2.5cm}|r}
    \toprule
    \midrule
    \textbf{Dataset} & \textbf{Description} & \textbf{Example} & \textbf{Ground-truth} & \textbf{Size}\\
    \midrule
    Physics & Answer the formula or the answer according to the question. & The Airy function $A_i(z)$ is famously defined by the integral \dots denotes the Cauchy principal value. What is the value of $I(a)$? & $\sin\!\left(\dfrac{a^{2}}{2}\right)$ & 17\,k\\
    \midrule
    Chemistry & Check whether the properties of the chemical molecules are correct. & BACE1 is an aspartic acid protease. Is this molecule effective in the experiment? Output ``True'' or ``False''. Molecule: [F][C][=C][C][=Branch1]\dots & True & 8\,k\\
    \midrule
    Mol2Caption & Generate the corresponding language description based on the molecular formula. & Molecule SELFIES:[C][N][/C] [=Branch2][Ring1][=N] [=C][/C][=C][C][=N$^+$] [Branch1][N][C][C][C][N$^+$] [Branch1][C][C][Branch1] [C][C][C][C][=C][C][=C] & The molecule is an unsymmetrical cationic C1 cyanine dye having . & 10\,k\\
    \midrule
    Caption2Mol & Generate the corresponding molecular formula based on the description. & The molecule is a member of the class of imidazoles that is 1H-imidazole in which the hydrogens at positions 2, 4, and 5 \dots, a member of pyridines and an compound. & [O][C][=C][C][=C] [Branch2][Ring1] [S][C][=N][C] [Branch1] & 10\,k\\
    \bottomrule
  \end{tabular}
\end{table}

\subsubsection{Scientific Reasoning Datasets}
We have constructed four sets of scientific reinforcement-learning data spanning the three disciplines of physics, chemistry, and biology. Below are brief descriptions of each. Detailed information about these datasets is summarized in Table~\ref{tab:rl_datasets_transposed}.
\begin{itemize}[noitemsep,nolistsep]
        \item \textbf{Physics Dataset:} consits of Ph.D.-level high-quality samples, encompassing a wide range of core mathematical disciplines such as mathematical analysis, linear algebra, probability theory, complex analysis, special functions, differential equations, group theory, topology, and differential geometry. It also integrates the latest applications of these fields in theoretical physics, making it a comprehensive and cutting-edge resource for advanced mathematical studies.
        \item  \textbf{Chemistry Dataset:} aims for molecular property prediction tasks, covering diverse properties such as drug toxicity, reactivity, and blood–brain barrier permeability. Its goal is to systematically enhance a model's accuracy and generalization ability in identifying molecular attributes.
        \item \textbf{Mol2Caption Dataset:} comprises high-quality chemical molecules paired with corresponding natural-language descriptions, designed to train large language models' ability to transform chemical structures into precise and detailed textual representations. The collection spans molecules from diverse chemical classes and functionalities, challenging models to generate semantically coherent yet structurally accurate and varied descriptions. It thoroughly enhances the model's comprehension of chemical language and its generative proficiency.
        \item  \textbf{Caption2Mol Dataset:} consits of natural-language descriptions paired with chemical molecules, intended to train large language models to generate molecular structures that satisfy a given textual specification. The descriptions encompass functional attributes, structural features, and application domains. Such a dataset emphasizes the model's accuracy in mapping text to molecular structures, especially its precision and diversity in converting natural-language descriptions into valid chemical structures.
\end{itemize} 

\subsubsection{General Reasoning Dataset}
To preserve general capabilities and prevent catastrophic forgetting during post-training after scientific continued pretraining, we curate a comprehensive general dataset spanning multiple domains for reinforcement learning.
\begin{itemize}[noitemsep,nolistsep]
        \item \textbf{Math:} We incorporate recent mathematical reasoning collections including DAPO~\cite{yu2025dapoopensourcellmreinforcement} and DeepScaler~\cite{deepscaler2025}, which compile numerous mathematical datasets featuring competition problems such as AIME~\cite{aime} and AMC~\cite{amc2023}. These datasets help in maintaining mathematical reasoning capabilities during post-training refinement.
        \item  \textbf{Stem:} We utilize WebInstruct-Verified~\cite{ma2025generalreasoneradvancingllmreasoning}, a dataset derived from web content and refined using LLMs, as the primary source for general science questions. To enable efficient reward verification during reinforcement learning, we focus on data with numeric, multiple-choice, or Boolean answers.
        \item \textbf{Code:} We collect programming problems from online coding platforms, programming competitions, and synthetic code generation tasks. Our collection incorporates real-world problems from LeetCode~\cite{xia2025leetcodedatasettemporaldatasetrobust} and curated problems from TACO-Verified~\cite{likaixin2024taco-verified}, ensuring robust general programming skills alongside scientific expertise.  
\end{itemize} 

\section{Upcycle Training}
\begin{figure}[t!]
    \centering
    \includegraphics[width=0.9\textwidth]{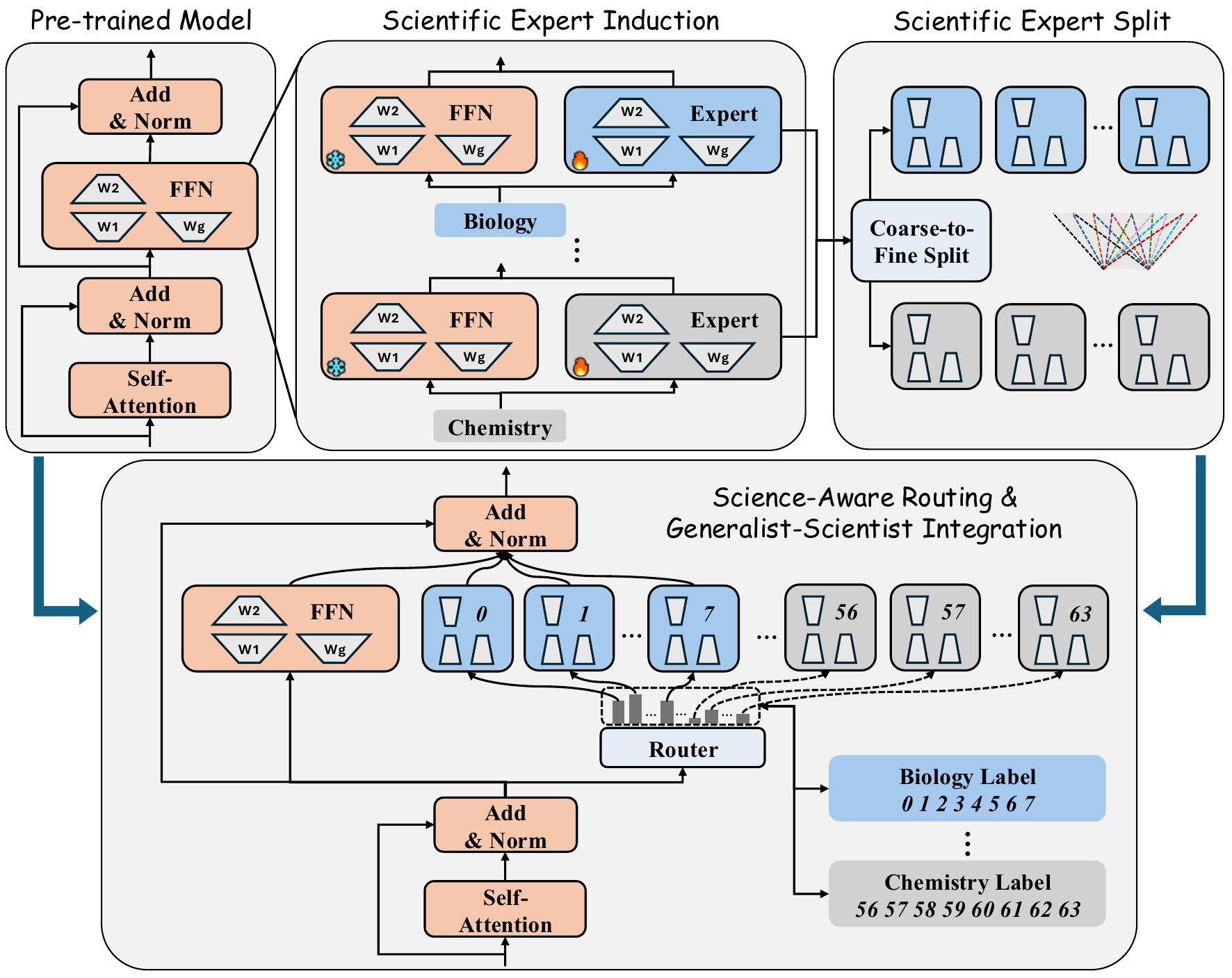}
    \caption{The framework of the Innovator. It is upcyle trained from the Qwen2.5-7B~\cite{Qwen2.5} dense model via the proposed novel four-stage training paradigm, including Scientific Expert Induction, Scientific Expert Split, Science-Aware Routing, and Generalist-Scientist Integration.
    }
    \label{fig:innovator_base}
\end{figure}

The proposed Innovator is trained via upcycling from the Qwen2.5-7B~\cite{Qwen2.5} dense model using 300B tokens. To substantially enhance scientific capabilities while preserving the general ability of the pre-trained LLM, we propose a novel four-stage training paradigm as shown in Fig.~\ref{fig:innovator_base}: 
\begin{itemize}
    \item \textbf{Scientific Expert Induction} preserves the generalization ability of the pre-trained dense model and inducts scientific experts;
    \item \textbf{Scientific Expert Split} divides the inducted experts corresponding to each single subject into multiple finer-grained ones;
    \item \textbf{Science-Aware Routing} guides the router to dispatch data from different subjects to the corresponding fine-grained experts;
    \item \textbf{Generalist-Scientist Integration} trains the entire model using hybrid data encompassing general and scientific knowledge for capability integration.
\end{itemize}

\subsection{Scientific Expert Induction} 
To preserve the general ability of the pre-trained LLM, which also lays the foundation for enhancing the scientific capability, we first conduct the induction training. Specifically, in each Feed-Forward Network (FFN) in the transformer block, we duplicate the FFN and get an expert for a single scientific subject. Then, given the training data from a single discipline, the activation from the prior attention block will be calculated by both the original FFN and the duplicated expert. It is noted that the pre-trained dense model is kept frozen, only the duplicated experts are optimized for scientific expert induction. 

For convenience, we omit the layer indices. Formally, denoting the input of each transformer block as $\boldsymbol{x}_{in}$, the forward calculation in the scientific expert induction process is modified as:
\begin{equation}
\begin{aligned}
\boldsymbol{x} &= \mathrm{LayerNorm}(\boldsymbol{x}_{in}), \\
\boldsymbol{x}_{attn} &= \mathrm{SelfAttn}(\boldsymbol{x}), \\
\boldsymbol{x}_{res} &= \boldsymbol{x}_{in} + \boldsymbol{x}_{attn}, \\
\boldsymbol{x} &= \mathrm{LayerNorm}(\boldsymbol{x}_{res}), \\
\boldsymbol{x} &= \mathrm{FFN}(\boldsymbol{x}) + \mathrm{Expert}(\boldsymbol{x}), \\
\boldsymbol{x} &= \boldsymbol{x}_{res} + \boldsymbol{x}, \\
\label{eq:exp_induc}
\end{aligned}
\end{equation}
in which $\mathrm{LayerNorm}$ denotes the layer normalization, $\mathrm{SelfAttn}$ is the self-attention calculation, $\mathrm{FFN}$ is the original FFN in the pre-trained LLM, $\mathrm{Expert}$ is duplicated from the original FFN and is  optimized with other modules kept frozen. By this way, we may induct biology expert using biology data only, and induct chemistry expert using chemistry data only, etc.

\subsection{Scientific Expert Split}
The inducted experts share the same architecture with the FFN in the pre-trained LLM, and the intermediate size $H$ is more than 5 times larger than the hidden size $h$ of the LLM. According to the analysis in DeepSeekMoE~\cite{dai2024deepseekmoe}, such architecture could cause significant redundancy among experts and limiting the flexible expert combination. Intuitively, solving complex scientific problems usually require multiple fine-grained capabilities for cooperation. Therefore, we propose to split the inducted coarse-grained scientific experts into multiple finer-grained ones.

Specifically, the inducted experts consist of 3 weight matrices: the up projection weight $W_1 \in \mathbb{R}^{h\times H}$, the gate weight $W_g \in \mathbb{R}^{h\times H}$, and the down projection weight $W_2 \in \mathbb{R}^{H\times h}$. Denoting the number of fine-grained experts that are split from an inducted expert as $N$, the weight matrices $W_1^i, W_g^i \in \mathbb{R}^{h\times \frac{H}{N}}$ and $W_2^i \in \mathbb{R}^{\frac{H}{N} \times h}$ in the $i$-th fine-grained expert is constructed as:
\begin{equation}
\begin{aligned}
W_1^i &= W_1[:, i \cdot \frac{H}{N} : (i+1) \cdot \frac{H}{N}], \\
W_g^i &= W_g[:, i \cdot \frac{H}{N} : (i+1) \cdot \frac{H}{N}], \\
W_2^i &= W_2[i \cdot \frac{H}{N} : (i+1) \cdot \frac{H}{N}, :], \\
i & \in [0, N-1].
\label{eq:exp_split}
\end{aligned}
\end{equation}

By this way, the inducted experts for each scientific subject will be split into multiple finer-grained experts with mitigated redundancy and improved combination diversity and flexibility, which lays the foundation for cross-disciplinary expert collaboration.

\subsection{Science-Aware Routing} 
Following the well-defined process of scientific expert segmentation, we can categorize scientific experts into distinct groups, each comprising specialists in a specific discipline. A router is essential to efficiently manage the allocation of data across these disciplines. To optimize the router’s performance and ensure accurate data dispatch, we implement a warmup training phase.

During this phase, each data sample is meticulously labeled based on its corresponding scientific discipline. This labeling is crucial for providing the router with the necessary information to make informed dispatch decisions. We then employ a multi-label classification loss function to train the router, as detailed in Eq.~\ref{eq:router_loss}.
\begin{equation}
\begin{aligned}
\hat{\boldsymbol{y}} &= \mathrm{softmax}(\mathrm{Router}(\boldsymbol{x})) \\
L_{cls} &= \sum_i \boldsymbol{y}_i\cdot \mathrm{log}(\hat{\boldsymbol{y}}_i),
\label{eq:router_loss}
\end{aligned}
\end{equation}
where $\boldsymbol{y}$ represents the label indicating which scientific experts $\boldsymbol{x}$ should be routed to. 
This loss function handles the complexities of multi-label classification tasks, enabling the router to learn from the labeled data and enhance its ability to accurately dispatch data to appropriate expert groups.
It is worth noting that the parameters of the attention layers and feed-forward networks are kept frozen during this process to preserve the expertise of each expert. We also adopt load balance loss combined with multi-label classification loss to encourage expert specialization.

\subsection{Generalist-Scientist Integration}
The development of the Innovator architecture involves two key processes: the induction and segmentation of scientific experts, and the implementation of science-aware routing. The first process generates multiple scientific experts, each specialized in distinct disciplines. The second process yields routers capable of directing data to the appropriate expert groups based on their scientific relevance.
By integrating a shared general expert with these multiple scientific experts and the router, the forward computation of each MoE block in Innovator can be formulated as:
\begin{equation}
\begin{aligned}
W &= \mathrm{softmax}(\mathrm{TopK}(\mathrm{Router}(\boldsymbol{x}))) \\
\boldsymbol{x} &= \mathrm{FFN}(\boldsymbol{x}) + \sum_{k}W_k\cdot \mathrm{Expert_k}(\boldsymbol{x}),
\label{eq:forward_innovator}
\end{aligned}
\end{equation}
where the calculation follows the topk-then-softmax order, and $\mathrm{TopK}$ selects the higher $K$ logits. $\mathrm{FFN}$ is the shared general expert initialized by the Qwen2.5-7B dense model, $\mathrm{Expert_k}$ is the $k$-th scientific expert, and $\mathrm{Router}$ is the science-aware router. For simplicity, we omit the layer indices and denote the input to each MoE block as $\boldsymbol{x}$. To integrate the general and scientific capabilities within the Innovator architecture, we perform continued pre-training on the Innovator model. During this phase, all parameters are jointly optimized using a dataset comprising 300 billion tokens, with each sequence consisting of 8192 tokens. This comprehensive training ensures that the model effectively combines general knowledge with domain-specific expertise. The entire upcylcle training is implemented based on the Megatron-LM~\cite{shoeybi2019megatron} framework for training efficiency and flexible parallelism, especially the expert parallelism.

\section{Post Training}
In the post-training stage, we primarily apply the Group Relative Policy Optimization (GRPO) algorithm~\cite{grpo2024} to perform reinforcement-learning fine-tuning on the Innovator, resulting in the Innovatore-Reason. For a specific question-answer pair $(q, a)$, the behavior policy $\pi_{\theta_{old}}$ samples a group of $G$ individual responses $\left \{ o_i \right \}_{i=1}^{G} $. Then, the advantage $\hat{A}_{i,t}$ of the $i$-th response is calculated by normalizing the group-level rewards $\left \{ R_i \right \}_{i=1}^{G}$:  

\begin{equation}
    \hat{A}_{i,t} = \frac{r_i-\text{mean}(R_i)_{i=1}^{G} }{\text{std}(\{ R_i\}_{i=1}^{G}) }.
\end{equation}
The GRPO objective updates model parameters as follows: 
\begin{equation}
\begin{split}
    \mathcal{J}_{\text{GRPO}}(\theta )&=\mathbb{E}_{(q,a)\sim \mathcal{D},\left \{ o_i \right \}_{i=1}^{G}\sim\pi_{\theta _{old}}(\cdot |q) } ,\\
    &=\biggl[ \frac{1}{G}\sum_{i=1}^{G}\frac{1}{|o_i|}\sum_{t=1}^{|o_i|} \biggl(\text{min} \Big(   r_{i,t}(\theta ) \hat{A}_{i,t}, \text{clip}(r_{i,t}(\theta ), 1-\varepsilon_{\text{low} },1+\varepsilon_{\text{high} })\hat{A}_{i,t} \Big) -\beta D_{\text{KL}}(\pi_{\theta}|| \pi_{\theta_{ref}})    \biggr)   \biggr],
\end{split}
\end{equation}

where 

\begin{equation}
    r_{i,t}(\theta)= \frac{\pi_{\theta}(o_{i,t}|q,o_{i,<t})}{\pi_{\theta_{old}}(o_{i,t}|q,o_{i,<t})}.
\end{equation}

Drawing on prior GRPO refinements, we adopt the DAPO~\cite{yu2025dapoopensourcellmreinforcement} configuration, including: 
(1) \textbf{Clip-Higher} to loosen the upper PPO-clip bound $\varepsilon _{high}$ to allow low-probability tokens to rise,
(2) \textbf{Dynamic Sampling} to adjust mini-batch composition on-the-fly,
(3) \textbf{Token-Level Loss} to compute per-token advantages for finer credit assignment, and
(4) \textbf{Overlong Reward Shaping} to penalize responses that exceed the length limit.

\section{Evaluation}
\subsection{General Evaluation} 
We conduct a comprehensive evaluation on general tasks using the widely-recognized Language Model Evaluation Harness~\cite{eval-harness} framework and the vLLM~\cite{vllm} inference serving engine. This evaluation benchmarks general performance across a spectrum of core capabilities, including natural language understanding, general question answering, mathematics reasoning and multilingual proficiency. The general evaluation datasets are categorized and listed as follows: 

\textbf{General Tasks.} MMLU~\citep{mmlu} (5-shot), BBH~\citep{bbh} (3-shot), ARC-Easy~\citep{arc} (25-shot), ARC-Challenge (25-shot), HellaSwag~\citep{hellaswag} (10-shot), WinoGrande~\citep{winogrande} (5-shot), NQ-Open~\citep{nqopen, nq} (5-shot), AGIEval~\citep{agieval} (0-shot), TriviaQA~\citep{triviaqa} (5-shot).

\textbf{Mathematics Tasks.} GPQA~\citep{gpqa} (5-shot), GSM8K~\citep{gsm8k} (8-shot), CMATH~\citep{cmath} (0-shot).

\textbf{Multilingual Tasks.} CLUEWSC2020~\citep{clue} (5-shot), C-Eval~\citep{ceval} (5-shot), CMMLU~\citep{cmmlu} (5-shot).

For the evaluation of base models, Innovator is compared with Qwen2.5-7B~\citep{Qwen2.5} and the Qwen2.5-7B-CT, which is continued trained from the Qwen2.5-7B under identical experimental settings.
\begin{table}[t!]
  \centering
  \small
  \caption{The performance comparison among the pre-trained Qwen2.5-7B, Qwen2.5-7B-CT, and Innovator on general tasks. Qwen2.5-7B-CT refers to the Qwen2.5-7B dense model continued trained on the 300B tokens adopted in the Innovator.}
  \label{tab:general_comp}
  \begin{tabular}{p{2.5cm}<{\centering} | p{2.5cm}<{\raggedright} | p{2.5cm}<{\centering} p{3cm}<{\centering} p{2.5cm}<{\centering} }
    \toprule
    \midrule
    \multicolumn{2}{c|}{\textbf{Benchmarks}} & \textbf{Qwen2.5-7B} & \textbf{Qwen2.5-7B-CT} & \textbf{Innovator} \\
    \midrule
    \multirow{9}{*}{General} & MMLU & 74.3 & 67.9 & 69.2 \\
    & BBH & 68.3 & 69.0 & 70.1 \\
    & ARC-Easy & 87.1 & 84.7 & 85.9 \\
    & ARC-Challenge & 63.6 & 60.3 & 61.9 \\
    & HellaSwag & 80.2 & 80.7 & 81.4 \\
    & WinoGrande & 76.3 & 76.1 & 77.1 \\
    & NQ-Open & 24.2 & 22.7 & 23.4 \\
    & AGIEval & 56.2 & 51.1 & 52.2 \\
    & TriviaQA & 60.0 & 53.2 & 54.6 \\
    \midrule
    \multirow{3}{*}{Math} & GPQA & 35.0 & 33.5 & 33.3 \\
    & GSM8K & 84.3 & 81.8 & 83.0 \\
    & CMATH & 69.8 & 74.6 & 79.1 \\
    \midrule
    \multirow{3}{*}{Multilingual} & CLUEWSC2020 & 52.5 & 51.3 & 52.6 \\
    & C-Eval & 81.9 & 79.2 & 81.1 \\
    & CMMLU & 82.8 & 80.1 & 80.5 \\
\midrule
    \multicolumn{2}{c|}{\textbf{Average}} & 66.4 & 64.4 & 65.7 \\
    \bottomrule
  \end{tabular}
\end{table}

The performance of general evaluation is reported in the Table~\ref{tab:general_comp}. Compared with the Qwen2.5-7B dense model, Innovator maintains the general capability with only 1\% loss after training on 300B tokens. Additionally, we also conduct continued pre-training on the Qwen2.5-7B dense model using the 300B tokens adopted by the Innovator. Continued training on the dense model achieves the lowest average general performance. Specifically, compared with the Innovator, the performance on CMath achieved by Qwen2.5-7B-CT is significantly decreased by 4.5 points. The comparisons comprehensively exhibit the effectiveness of Innovator in retaining the general capabilities in scientific upcycle training.

\subsection{Scientific Evaluation}
To comprehensively evaluate the scientific performance of base LLMs, we collect diverse scientific tasks from the SciAssess benchmark~\citep{cai2024sciassessbenchmarkingllmproficiency} and the OlympicArena benchmark~\citep{huang2024olympicarena}. SciAssess~\citep{cai2024sciassessbenchmarkingllmproficiency} is constructed mainly for evaluating the capability in scientific literature analysis. It contains a total of 6,888 questions across four major domains: biology, chemistry, materials, and medicine. The difficulty ranges from memorizing, comprehending and analytical reasoning the literature. Tasks in the SciAssess are various, such as multiple choice, text and table extraction. The OlympicArena~\citep{huang2024olympicarena} focuses on evaluating the multi-discipline cognitive reasoning capabilities. It includes 11,163 challenging bilingual problems spanning disciplines such as mathematics, physics, chemistry, biology, geography, astronomy, etc. For an efficient evaluation, we use vLLM~\citep{vllm} as the inference engine. 
\begin{table}[t!]
  \centering
  \small
  \caption{The scientific performance comparison among the pre-trained Qwen2.5-7B, Qwen2.5-14B, and Innovator. Science subjects include biology, chemistry, material, medicine, geography and astronomy.}
  \label{tab:sci_comp}
  \begin{tabular}{p{5cm}<{\centering} | p{2.5cm}<{\centering} p{2.5cm}<{\centering} p{2.5cm}<{\centering} }
    \toprule
    \midrule
    \textbf{Benchmarks} & \textbf{Qwen2.5-7B} & \textbf{Qwen2.5-14B} & \textbf{Innovator} \\
    \midrule
    MMLU-Pro-Biology & 45.5 & 59.5 & \textbf{69.9} \\
    Biology Chart QA & 31.7 & 39.7 & \textbf{39.7} \\
    Chemical Entities Recognition & 36.3 & 54.7 & \textbf{60.9} \\
    Compound Disease Recognition & 44.9 & 40.0 & \textbf{74.6} \\
    Disease Entities Recognition & 46.0 & 55.2 & \textbf{69.7} \\
    Gene Disease Function & 16.8 & 12.8 & \textbf{43.4} \\
    MMLU-Pro-Chemistry & 28.4 & 28.2 & \textbf{36.0} \\
    OLED Property Extraction & 0.4 & \textbf{28.7} & 2.0 \\
    Polymer Chart QA & 66.7 & 53.3 & \textbf{66.7} \\
    Polymer Composition QA & 25.5 & \textbf{41.8} & 32.7 \\
    Polymer Property Extraction & 6.0 & 14.3 & \textbf{22.4} \\
    Solubility Extraction & 23.5 & 24.7 & \textbf{25.6} \\
    Reactant QA & \textbf{22.1} & 20.5 & 20.5 \\
    Reaction Mechanism QA & 18.2 & 18.2 & \textbf{22.7} \\
    Material QA & 54.0 & \textbf{63.9} & 55.5 \\
    Alloy Chart QA & 53.3 & 60.0 & \textbf{60.0} \\
    Composition Extraction & \textbf{7.8} & 7.4 & 6.8 \\
    Temperature QA & 24.6 & \textbf{25.6} & 23.7 \\
    Sample Differentiation & \textbf{20.7} & 20.3 & 20.3 \\
    MMLU-Pro-Health & 36.5 & 43.9 & \textbf{55.5} \\
    Affinity Extraction & 1.2 & 3.2 & \textbf{3.9} \\
    Drug Chart QA & 40.0 & 40.0 & \textbf{46.7} \\
    Markush to Molecule & 23.9 & \textbf{35.3} & 20.4 \\
    Molecule in Document & \textbf{48.0} & 42.0 & 44.0 \\
    OlympicArena-Math & 17.0 & 9.7 & \textbf{21.8} \\
    OlympicArena-Physics & 17.9 & 10.8 & \textbf{19.1} \\
    OlympicArena-Chemistry & 30.6 & 18.8 & \textbf{32.3} \\
    OlympicArena-Biology & 35.5 & 27.3 & \textbf{36.9} \\
    OlympicArena-Geography & 34.8 & 30.5 & \textbf{37.5} \\
    OlympicArena-Astronomy & 24.6 & 13.8 & \textbf{28.4} \\
    \midrule
    \textbf{Average} & 29.4 & 31.5 & \textbf{36.6} \\
    \textbf{Win Rate} & 4/30 & 5/30 & \textbf{21/30} \\
    \bottomrule
  \end{tabular}
\end{table}

The scientific performance of the Innovator and the pre-trained Qwen2.5-7B/14B is compared in the Table~\ref{tab:sci_comp}. Within 30 scientific evaluation tasks, Innovator achieves the highest score on 21 tasks. Compared with Qwen2.5-7B, which the Innovator is extended from, our model achieves a significant performance improvement by 25\%. Compared with Qwen2.5-14B, even with slightly fewer activated parameters of 13.3B, the Innovator also exhibits great performance advantage by 16\%. 

Based on the evaluation of general and scientific performance, the effectiveness of the Innovator in preserving general capabilities and significantly improving scientific capabilities through expert expansion with 300B tokens has been systematically verified.

\subsection{Architecture Effectiveness Validation} 
\label{sec:arch_effec_val}
To verify the effectiveness of the architecture of the Innovator with keeping one shared expert without change, we compare the performance under different settings of the intermediate size and number of scientific experts. For a quick validation, the ablation studies are performed by training with 50B tokens using the same data mixture strategy of the Innovator. The average performance of the general and scientific evaluation are presented in Table~\ref{tab:arch_effec}.

According to the results, Innovator achieves the best performance on both general and scientific evaluation. Compared with the ablation settings, the intermediate size of the scientific experts in Innovator is the lowest, which contributes to the less redundancy among experts. Based on that, under a fixed parameter budget, finer-grained experts could result in a larger number of experts, thereby enhancing the flexibility and diversity of expert combinations. According to the results, the effectiveness of the architecture of the Innovator is comprehensively verified.

\subsection{Performance with Training Data Scaling Up} 
\begin{figure}[t!]
    \centering
    \includegraphics[width=0.8\textwidth]{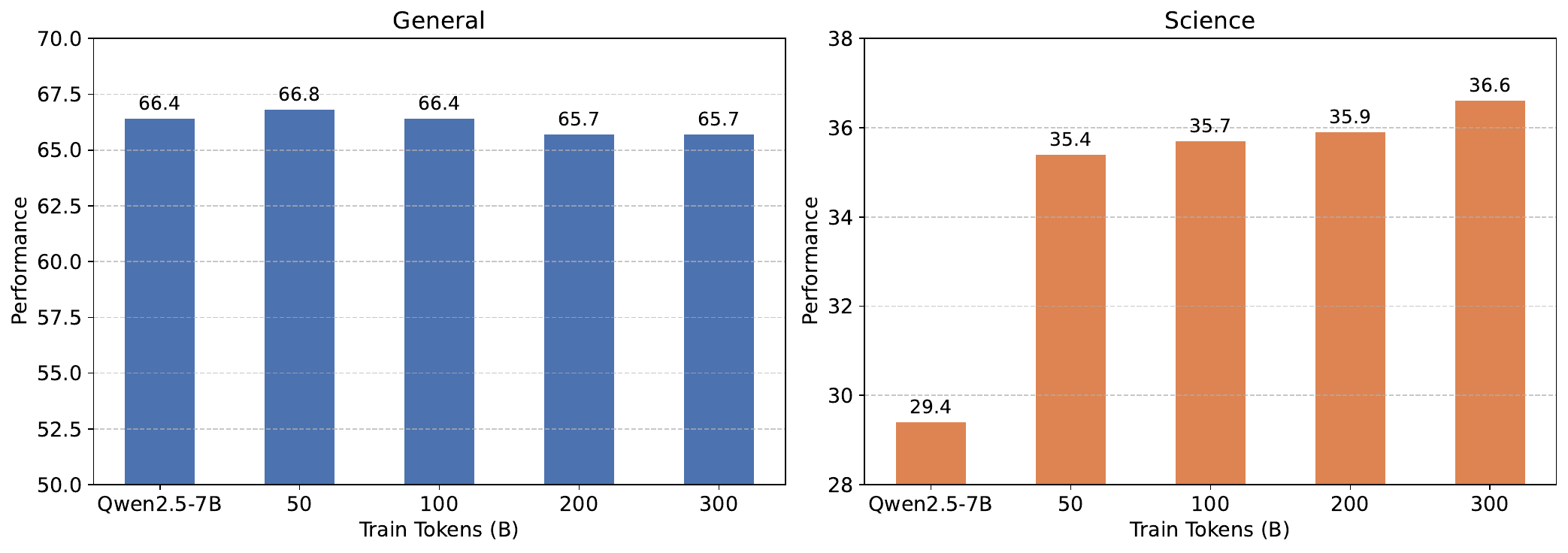}
    \caption{The general and scientific performance of the Innovator with training data scaling up.}
    \label{fig:data_scale}
\end{figure}

Except for the comparison across different architecture settings delivered in the Sec.~\ref{sec:arch_effec_val}, another perspective to prove the effectiveness of the Innovator is the performance with training data scaling up. To this end, based on the same data mixture strategy, we perform upcycle training using 50B, 100B, 200B, and 300B tokens, respectively. The general and scientific performance are compared in the Fig.~\ref{fig:data_scale}.
\begin{table}[t!]
  \centering
  \small
  \caption{The general and scientific performance comparisons across different settings of the intermediate size and number of scientific experts. All experiments are performed by training on 50B tokens.}
  \label{tab:arch_effec}
  \begin{tabular}{p{4cm}<{\centering} | p{1.5cm}<{\centering}  p{1.5cm}<{\centering} p{1.5cm}<{\centering} | p{3cm}<{\centering} }
    \toprule
    \midrule
    \textbf{Settings} & \multicolumn{3}{c|}{\textbf{Ablations}} & \textbf{Innovator} \\
    \midrule
    Intermediate Size & 18944 & 9472 & 4736 & 2368 \\
    Total Experts & 8 & 4 & 16 & 64 \\  
    Activated Experts & 1 & 2 & 4 & 8 \\
    \midrule
    \textbf{General Performance} & 66.7 & 66.6 & 66.6 & \textbf{66.8} \\
    \textbf{Scientific Performance} & 34.3 & 34.2 & 34.8 & \textbf{35.4} \\
    \bottomrule
  \end{tabular}
\end{table}

After upcycle training with 50B tokens, both general and scientific performance are improved compared with the Qwen2.5-7B. On the one side, the self-curated data is proved to be high-quality, especially in the collection of scientific knowledge. On the other side, the effectiveness of the model design in preserving general capability, as well as strengthening scientific ability is validated. By scaling up the training data, the general performance of the model remains close to the original performance, with a maximum loss of 1\%. Meanwhile, the scientific performance is constantly improving, and the best performance that surpasses Qwen2.5-7B by 25\% is achieved via training on 300B tokens. It further verifies the effectiveness of the Innovator in the generalist-scientist integration.

\subsection{Reasoning Evaluation} 
\begin{table}[t!]
  \centering
  \small
  \caption{The performance comparison among the pre-trained Qwen2.5-7B, Qwen2.5-7B-Instruct, Qwen2.5-14B, and Innovator-Reason on general, math and scientific reasoning tasks. Innovator-Reason denotes the model achieved by post-training on Innovator with reinforcement learning.}
  \label{tab:reason_rl}
  \begin{tabular}{p{1cm}<{\centering} | p{1.8cm}<{\raggedright} | p{2cm}<{\centering} p{3.5cm}<{\centering} p{2.5cm}<{\centering} p{3.0cm}<{\centering} }
    \toprule
    \midrule
    \multicolumn{2}{c|}{\textbf{Benchmarks}} & \textbf{Qwen2.5-7B} & \textbf{Qwen2.5-7B-Instruct} & \textbf{Qwen2.5-14B} & \textbf{Innovator-Reason} \\
    \midrule
    \multirow{2}{*}{General} & MMLU & 67.0 & 74.3 & 75.4 & \textbf{76.4} \\
    & MMLU-pro & 42.6 & 56.3 & 55.3 & \textbf{56.7} \\
    \midrule
    \multirow{6}{*}{Math} & AIME24 & 6.7 & \textbf{19.7} & 9.9 & 16.6 \\
    & AMC23 & 54.2 & 50.0 & 49.2 & \textbf{72.5} \\
    & Math-500 & 69.2 & 76.8 & 77.8 & \textbf{80.9} \\
    & OlymicBench  & 34.8 & 39.0 & 39.5 & \textbf{48.2} \\
    & GSM8K  & 88.6 & 91.8 & \textbf{94.2} & 92.7 \\
    & CMath  & 78.4 & \textbf{93.3} & 86.9 & 91.7 \\
    & MinervaMath  & 22.8 & 36.0 & 27.7 & \textbf{36.4} \\
    \midrule
    \multicolumn{2}{c|}{\textbf{Average}} & 51.6 & 59.7 & 57.3 & \textbf{63.6} \\
    \midrule
    \multirow{4}{*}{Science} & Physics & 11.5 & 12.8 & 11.4 & \textbf{18.3} \\
    & Chemistry & 45.2 & 64.6 & 63.4 & \textbf{84.5} \\
    & Mol2caption & 10.9 & \textbf{10.9} & 4.4 & 10.3 \\
    & Caption2mol & 10.7 & 8.6 & 13.5 & \textbf{15.5} \\
    \midrule
    \multicolumn{2}{c|}{\textbf{Average}} & 19.6 & 24.2 & 23.2 & \textbf{32.2} \\
    \bottomrule
  \end{tabular}
\end{table}

We evaluate the model post-trained on Innovator, namely Innovator-Reason, on a broad collection of general, mathematical, and scientific benchmarks. In the general domain, we use MMLU~\citep{mmlu} and MMLU-Pro~\cite{mmlu-pro} for evaluation. In the mathematical evaluation, we benchmark on AIME24~\cite{aime}, AMC23~\cite{amc2023}, MATH-500~\cite{math500}, OlymicBench~\cite{he2024olympiadbench}, GSM8K~\citep{gsm8k}, CMath~\citep{cmath}, and Minervamath~\cite{lewkowycz2022solvingquantitativereasoningproblems}. Additionally, we construct four test sets covering Physics, Chemistry, Mol2Caption, and Caption2Mol for scientific evaluation.

The performance of the Innovator-Reason on general, math and scientific tasks is reported in the Table~\ref{tab:reason_rl}. Compared with Qwen2.5-7B, Qwen2.5-7B-Instruct, and Qwen2.5-14B, Innovator-Reason delivers substantial improvements across all reasoning benchmarks. Concretely, in the non-scientific evaluation, Innovator-Reason surpasses Qwen2.5-7B, Qwen2.5-7B-Instruct, Qwen2.5-14B by around 23\%, 7\%, 11\%. Moreover, in the scientific evaluation, Innovator-Reason exhibits a much larger performance advantage, achieving the improvement by 64\%, 33\%, 39\% compared with Qwen2.5-7B, Qwen2.5-7B-Instruct, and Qwen2.5-14B, respectively. Specifically, in the field of chemistry, Innovator-Reason achieves the significant performance advantage by 87\% compared with Qwen2.5-7B, 31\% compared with Qwen2.5-7B-Instruct, and 33\% compared with Qwen2.5-14B. Furthermore, for an intuitive qualitative evaluation, we present the detailed process of Innovator-Reason in solving complex scientific problems from Table~\ref{tab:rl_good_case_physics} to Table~\ref{tab:rl_good_case_caption2mol} in the Appendix. Innovator-Reason produces logical and concise reasoning processes. The above results demonstrate that Innovator lays the solid foundation for reasoning boosting. Then, post-training can effectively elicit the scientific knowledge embedded in the Innovator, and further strengthen the model’s reasoning capabilities.

\section{Conclusion} 
Towards building a large language model with integrated general and scientific capabilities, we propose the Innovator, an MoE model with 53.3B total parameters and 13.3B activated parameters. Each MoE layer consists of 1 shared general expert and 64 fine-grained scientific experts with 8 activated. Innovator is upcycle trained from the Qwen2.5-7B dense model. To retain the general capability and enhance the scientific ability, we propose a robust data processing paradigm and a novel four-stage upcycle training paradigm. The data processing paradigm utilizes a multi-level quality control system for general pre-training data, an intelligent cleaning and restoration system for scientific literature, and a two-stage synthesis method for scientific instruction fine-tuning, ensuring high-quality and domain-specific training data. The four-stage upcycle training paradigm firstly inducts the experts using data from specific disciplines, then divides the inducted experts into multiple finer-grained ones for enhancing the diversity and flexibility of expert combinations. After science-aware routing for accurate data dispatch, the final stage trains the entire model for generalist-scientist integration. Compared with Qwen2.5-7B, Innovator preserves 99\% general capability, while significantly improves scientific capability by 25\%, and even surpasses that of Qwen2.5-14B by 16\%. Furthermore, we conducted reinforcement learning based on Innovator using multiple general and scientific reasoning datasets, resulting in the Innovator-Reason. Results show that Innovator-Reason achieves significant improvements in scientific reasoning by over 30\%, exhibiting its strong reasoning capability for solving complicated scientific problems.

\bibliographystyle{unsrt}
\bibliography{main}

\newpage
\appendix
\section{Contributions and Acknowledgments} 
We acknowledge all contributors of this work, they are affilated to: $^{1}$Shanghai Jiao Tong University; $^{2}$MemTensor (Shanghai) Technology Co., Ltd.; $^{3}$Institute for Advanced Algorithms Research, Shanghai; $^{4}$Institute of Physics CAS; $^{5}$DP Technology.

Contributors are listed below:

\textbf{Core Contributors:} Ning Liao$^{1,2}$, Xiaoxing Wang$^{1,2}$.

\textbf{Upcycle Training:} Ning Liao$^{1,2}$, Xiaoxing Wang$^{1,2}$, Feng Hong$^{1}$, Geng Yu$^{1}$, Zihua Zhao$^{1}$.

\textbf{Data Preparation:} Zhiyu Li$^{2,3}$, Zehao Lin$^{2,3}$, Shixiang Song$^{1,2}$, Kai Chen$^{2}$.

\textbf{General Evaluation:} Xiaoxing Wang$^{1,2}$, Geng Yu$^{1}$, Xiangqi Jin$^{1}$, Xiaohan Qin$^{1}$.

\textbf{Scientific Evaluation:} Ning Liao$^{1,2}$, Zihua Zhao$^{1}$, Sitao Xie$^{1}$, Longxuan Wei$^{1}$.

\textbf{Post Training:} Ning Liao$^{1,2}$, Weiyang Guo$^{1}$, Feng Hong$^{1}$, Xiangqi Jin$^{1}$, Jiale Ma$^{1}$.

\textbf{Scientific Reasoning Benchmark:} Kun Chen$^{4}$, Hengxing Cai$^{5}$, Guojiang Zhao$^{5}$.

\textbf{Advising:} Linfeng Zhang$^{1}$, Yanfeng Wang$^{1}$, Feiyu Xiong$^{2,3}$, Jiangchao Yao$^{1}$, Zhouhan Lin$^{1}$, Junchi Yan$^{1}$.

\section{Appendix}
The qualitative examples showing the detailed process of the proposed Innovator-Reason in solving Physics, Chemistry, Mol2Caption and Caption2Mol problems are delivered from Table~\ref{tab:rl_good_case_physics} to Table~\ref{tab:rl_good_case_caption2mol}. Innovator-Reason produces logical and concise reasoning process in solving problems, showing its outperforming capability in scientific deep thinking.
\begin{table}[h]
  \centering
  \small
  \caption{The process of Innovator-Reason in solving the Physics problem.}
  \label{tab:rl_good_case_physics}
  \begin{tabular}{p{3cm} | p{12.5cm} }
    \toprule
    \midrule

    \textbf{User} & 
    You are a helpful assistant. 
    
    \vspace{4pt}
    
    An infinitely long, hollow cylinder is defined by an inner radius $a$ and an outer radius $b$. The cylinder is non-conducting and carries a static, non-uniform volume charge density $\rho(r)$ that depends only on the radial distance $r$ from the central axis. The charge density is given by $
    \rho(r) = \begin{cases} k r^{-3/2} & \text{for } a \le r \le b \\ 0 & \text{otherwise} \end{cases}$, where $k$ is a positive constant. The relationship between the inner and outer radii is given by $b=5a$.
    
    \vspace{4pt}

    Your task is to find the radial distance $r_{\text{max}}$ from the axis at which the magnitude of the electric field is maximum. Express your answer in terms of the inner radius $a$.
    
    \\\midrule

    \textbf{Innovator-Reason Thinking Process} & To find the radial distance $r_{\text{max}}$ at which the magnitude of the electric field is maximum, we need to follow these steps:

    \vspace{4pt}
    
    \textbf{1. Calculate the total charge $Q$ within the cylinder:} The charge density is given by $\rho(r) = k r^{-3/2}$ for $a \le r \le b$. The total charge $Q$ can be found by integrating the charge density over the volume of the cylinder:
    $Q = \int_{a}^{b} \rho(r) \cdot 2\pi r \, dr = \int_{a}^{b} k r^{-3/2} \cdot 2\pi r \, dr = 2\pi k \int_{a}^{b} r^{-1/2} \, dr
    $, $Q = 2\pi k \left[ 2r^{1/2} \right]_{a}^{b} = 4\pi k \left( b^{1/2} - a^{1/2} \right)$

    \vspace{4pt}
    
    \textbf{2. Use Gauss's Law to find the electric field $E$:} Gauss's Law states that the flux of the electric field through a closed surface is equal to the charge enclosed divided by $\epsilon_0$. For a cylindrical surface at radius $r$, the electric field $E$ is constant and perpendicular to the surface, so: $E \cdot 2\pi r = \frac{Q_{\text{enc}}}{\epsilon_0}$, where $Q_{\text{enc}}$ is the charge enclosed within radius $r$. The enclosed charge is: $
    Q_{\text{enc}} = \int_{a}^{r} k r'^{-3/2} \cdot 2\pi r' \, dr' = 2\pi k \int_{a}^{r} r'^{-1/2} \, dr' = 4\pi k \left( r^{1/2} - a^{1/2} \right)$. Thus, the electric field is: $
    E = \frac{4\pi k \left( r^{1/2} - a^{1/2} \right)}{2\pi r \epsilon_0} = \frac{2k \left( r^{1/2} - a^{1/2} \right)}{r \epsilon_0}$.
    
    \vspace{4pt}
    
    \textbf{3. Find the maximum of $E$ with respect to $r$:} To find the maximum, we take the derivative of $E$ with respect to $r$ and set it to zero: $
    \frac{dE}{dr} = \frac{2k}{\epsilon_0} \left( \frac{1}{2} r^{-1/2} - \frac{r^{1/2} - a^{1/2}}{r^2} \right) = 0$. Simplifying, we get:
    $\frac{1}{2} r^{-1/2} = \frac{r^{1/2} - a^{1/2}}{r^2}$, $r^{3/2} = 2r^{1/2} - 2a^{1/2}$, $r^{3/2} - 2r^{1/2} + 2a^{1/2} = 0$. Let $x = r^{1/2}$, then: $x^3 - 2x + 2a^{1/2} = 0$, Solving this cubic equation, we find that the maximum occurs at: $
    r_{\text{max}} = 4a$.
    
    Thus, the radial distance at which the magnitude of the electric field is maximum is \textbf{4a}.\\\midrule

    \textbf{Innovator-Reason Answer} & 4a.\\
    
    \bottomrule
  \end{tabular}
\end{table}

\begin{table}[t!]
  \centering
  \small
  \caption{The process of Innovator-Reason in solving the Chemistry problem.}
  \label{tab:rl_good_case_chemistry}
  \begin{tabular}{p{3cm} | p{13.5cm} }
    \toprule
    \midrule
    \textbf{User} & 
    \textbf{Role:} You are a top AI assistant specializing in molecular chemistry and drug discovery, proficient in molecular property prediction.
    
    \vspace{4pt}

    \textbf{Task:} Evaluate whether the specified molecule is mutagenic. Mutagenicity means the ability of a drug to induce genetic alterations. The Ames test is a short-term bacterial reverse mutation assay detecting a large number of compounds which can induce genetic damage and frameshift mutations. 
    Output "True" or "False".
    
    \vspace{4pt}
    
    \textbf{Molecule to analyze:} $\texttt{[C][C][C][C][C@@H1][Branch1][Ring1][C][C][C][O][C]}$
    
    $\texttt{[=Branch1][C][=O][C][C][C][C][C][C][C][C][C][=Branch1][C][=O][O][C][C@H1]]}$
    
    $\texttt{[Branch1][Ring1[C][C][C][C][C][C]}$. Please reason step by step.

    \\\midrule

    \textbf{Innovator-Reason Thinking Process} & To determine whether the specified molecule is mutagenic, we need to analyze its structure and compare it with known mutagenic compounds. Here's a step-by-step breakdown of the process:

    \vspace{4pt}
    
    \textbf{1. Identify the Molecule Structure:} The given molecule is: $\texttt{[C][C][C][C][C@@H1][Branch1]}$
    
    $\texttt{[Ring1][C][C][C][O][C][=Branch1][C][=O][C][C][C][C][C][C][C][C][C][=Branch1][C]}$
    
    $\texttt{[=O][O][C][C@H1][Branch1][Ring1][C][C][C][C][C][C]}$.

    \vspace{4pt}
    
    \textbf{2. Analyze the Structure:} The molecule contains a branched alkyl chain with a double bond. It includes a ring structure (Ring1) and a branch (Branch1). There are oxygen atoms present, which can form hydrogen bonds and potentially affect the molecule's reactivity.
    
    \vspace{4pt}
    
    \textbf{3. Compare with Known Mutagenic Compounds}: Mutagenic compounds often contain specific functional groups that can interact with DNA, such as alkylating agents, aromatic amines, or other reactive species. The presence of a double bond and a ring structure does not inherently make a molecule mutagenic, but it can contribute to reactivity.
    
    \vspace{4pt}
    
    \textbf{4. Consider the Ames Test}: The Ames test is a short-term bacterial reverse mutation assay used to detect compounds that can induce genetic damage. Compounds that are mutagenic in the Ames test often have specific structural features that allow them to interact with DNA.
    
    \vspace{4pt}
    
    \textbf{5. Conclusion}: Based on the given structure, there are no clear mutagenic features such as alkylating agents or aromatic amines. The presence of a double bond and a ring structure does not necessarily indicate mutagenicity.

    Given the lack of specific mutagenic features in the molecule, the conclusion is: \textbf{False}.
    \\\midrule
    
    \textbf{Innovator-Reason Answer} & False. \\
    
    \bottomrule
  \end{tabular}
\end{table}

\begin{table}[t!]
  \centering
  \small
  \caption{The process of Innovator-Reason in solving the Mol2Caption problem.}
  \label{tab:rl_good_case_mol2caption}
  \begin{tabular}{p{3cm} | p{13.5cm} }
    \toprule
    \midrule
    \textbf{User} & 
    \textbf{Role:} You are a helpful assistant. You are a professional chemist.
    
    \vspace{4pt}

    \textbf{Task:} Your task is to generate a natural, concise, and chemically accurate description of a molecule based on its SELFIES representation.
    
    \vspace{4pt}
    
    \textbf{Example (Molecular SELFIES):} 
    
    \quad $\texttt{[C][C][Branch1][C][C][Branch2][Branch1][=Branch2][C][O][P][=Branch1][C][=O]}$
    
    \quad $\texttt{[Branch1][C][O][O][P][=Branch1][C][=O][Branch1][C][O][O][C][C@H1][O][C@@H1]}$
    
    \quad $\texttt{[Branch1][P][N][C][=N][C][=C][Branch1][C][N][N][=C][N][=C][Ring1][\#Branch1]}$
    
    \quad $\texttt{[Ring1][\#Branch2][C@H1][Branch1][C][O][C@@H1][Ring1][S][O][P][=Branch1][C]}$
    
    \quad $\texttt{[=O][Branch1][C][O][O][C@@H1][Branch1][C][O][C][=Branch1][C][=O][N][C][C][C]}$
    
    \quad $\texttt{[=Branch1][C][=O][N][C][C][S][C][=Branch1][C][=O][C][/C][=C][\textbackslash C][C][=O]}$

    \vspace{4pt}
    
    \textbf{Example (Answer):} The molecule is a cis-3-enoyl-CoA and a monounsaturated fatty acyl-CoA. It derives from an adipoyl-CoA. It is a conjugate acid of a cis-3,4-didehydroadipoyl-CoA semialdehyde(4-)

    \vspace{4pt}

    \textbf{Problem:} Now try this. \textbf{Molecule SELFIES}: $\texttt{[C][=C][Branch2][=Branch1][=Branch1]}$
    
    \quad $\texttt{[O][C@H1][C@H1][Branch1][C][O][C@@H1][Branch1][Ring1][C][O][O][C][Branch2]}$
    
    \quad $\texttt{[Ring2][N][O][P][=Branch1][C][=O][Branch1][C][O-1][O][P][=Branch1][C][=O]}$
    
    \quad $\texttt{[Branch1][C][O-1][O][C][C@H1][O][C@@H1][Branch1][=N][N][C][=C][C][=Branch1][C]}$
    
    \quad $\texttt{[=O][NH1][C][Ring1][\#Branch1][=O][C@H1][Branch1][C][O][C@@H1][Ring1][=C][O]}$
    
    \quad $\texttt{[C@@H1][Ring2][Ring2][C][N][C][Branch1][C][C][=O][C][=Branch1][C][=O][O-1]}$ 

    \vspace{4pt}
    
    Think step by step.
    
    \\\midrule

    \textbf{Innovator-Reason Thinking Process} & To generate a natural, concise, and chemically accurate description of the given molecule based on its SELFIES representation, I will follow these steps:

    \vspace{4pt}
    
    \textbf{Identify the components and structure of the molecule.} 
    
    \vspace{4pt}

    1) The molecule contains several parts such as carbon chains, oxygen atoms, a nitrogen atom, and a sulfur atom. 2) There are multiple branches and rings in the structure.
    
    \vspace{4pt}
    
    \textbf{Analyze the functional groups and bonds}

    \vspace{4pt}
    
    1) There are carboxyl groups, aldehyde groups, and other polar groups. 2) There are also oxygen atoms that are part of aldehyde groups and other oxygen atoms that are part of other functional groups.

    \vspace{4pt}
    
    \textbf{Determine the overall structure and the specific compounds it could be}
    
    \vspace{4pt}
    
    1) The molecule contains a conjugate base of a carboxylic acid. 2) There are multiple aldehyde groups and a hydroxyl group. 3) There are also parts of the structure that could be part of a fatty acid or a similar compound.
    
    \vspace{4pt}
    
    \textbf{Identify the specific compound}
    
    \vspace{4pt}
    
    1) Given the components and the structure, the molecule could be a specific type of fatty acid or a derivative of a fatty acid. 2) It could also be a conjugate base of a specific acid.
    
    \vspace{4pt}
    
    \textbf{Formulate the final description}

    \vspace{4pt}
    
    The molecule is an epoxy(hydroxy)icosatrienoate that is the conjugate base of 11-hydroxy-(14R,15S)-epoxy-(5Z,8Z,12E)-icosatrienoic acid, obtained by deprotonation of the carboxy group; major species at pH 7.3. It is a conjugate base of an 11-hydroxy-(14R,15S)-epoxy-(5Z,8Z,12E)-icosatrienoic acid.
    
    \vspace{4pt}
    
    \textbf{Final Answer:} The molecule is an epoxy(hydroxy)icosatrienoate that is the conjugate base of 11-hydroxy-(14R,15S)-epoxy-(5Z,8Z,12E)-icosatrienoic acid, obtained by deprotonation of the carboxy group; major species at pH 7.3. It is a conjugate base of an 11-hydroxy-(14R,15S)-epoxy-(5Z,8Z,12E)-icosatrienoic acid.
    
    \\\midrule

    \textbf{Innovator-Reason Answer} & 11-hydroxy-(14R,15S)-epoxy-(5Z,8Z,12E)-icosatrienoic acid, obtained by deprotonation of the carboxy group; major species at pH 7.3. It is a conjugate base of an 11-hydroxy-(14R,15S)-epoxy-(5Z,8Z,12E)-icosatrienoic acid.\\
    
    \bottomrule
  \end{tabular}
\end{table}

\begin{table}[t!]
  \centering
  \small
  \caption{The process of Innovator-Reason in solving the Caption2Mol problem.}
  \label{tab:rl_good_case_caption2mol}
  \begin{tabular}{p{3cm} | p{13.5cm} }
    \toprule
    \midrule
    \textbf{User} & 
    \textbf{Role:} You are a helpful assistant.
    
    \vspace{4pt}

    \textbf{Task:} Please generate the molecular structure (SELFIES) based on the following description.
    
    \vspace{4pt}
    
    \textbf{Example (Description):} The molecule is a benzoate ester obtained by the formal condensation of the carboxy group of 4-[(2,4-dihydroxy-6-propylbenzoyl)oxy]-2-hydroxy-6-propylbenzoic acid with the 4-hydroxy group of 2,4-dihydroxy-6-pentylbenzoic acid (olivetolic acid). It is isolated from the endophytic fungi Cytonaema and acts as an inhibitor of human cytomegalovirus (hCMV) protease. It has a role as a metabolite, an antiviral agent and a protease inhibitor. It is a benzoate ester, a member of resorcinols and a monohydroxybenzoic acid. It derives from an olivetolic acid.

    \vspace{4pt}
    
    \textbf{Example (Answer):} $\texttt{[C][C][C][C][C][C][=C][C][Branch2][Ring2][=C][O][C][=Branch1]}$
    
    \quad $\texttt{[C][=O][C][=C][Branch1][C][O][C][=C][Branch2][Ring1][\#Branch1][O][C][=Branch1]}$
    
    \quad $\texttt{[C][=O][C][=C][Branch1][C][O][C][=C][Branch1][C][O][C][=C][Ring1][Branch2]}$
    
    \quad $\texttt{[C][C][C][C][=C][Ring2][Ring1][Branch1][C][C][C][=C][C][Branch1][C][O][=C]}$
    
    \quad $\texttt{[Ring2][Ring2][C][C][=Branch1][C][=O][O]}$

    \vspace{4pt}

    \textbf{Problem:} Now try this. \textbf{Description}: The molecule is a carbamate ester of thymidine in which the 5'-hydroxy group has been esterified with [3-(benzoyloxy)-3-oxopropyl]carbamic acid, It derives from a thymidine. Please think step by step and return the final answer.
    
    \\\midrule

    \textbf{Innovator-Reason Thinking Process} & To generate the molecular structure (SELFIES) for the given description, I will follow these steps:
    
    \vspace{4pt}
    
    \textbf{Step 1: Identify the Key Components and Bonds}
    
    \vspace{4pt}
    
    The molecule is a carbamate ester of thymidine with the 5'-hydroxy group esterified with [3-(benzoyloxy)-3-oxopropyl]carbamic acid. This means the structure involves a thymidine backbone with a specific ester and carbamate group attached.
    
    \vspace{4pt}
    
    \textbf{Step 2: Break Down the Components}

    \vspace{4pt}
    
    1) The base component is thymidine. 2) The ester group is [3-(benzoyloxy)-3-oxopropyl]carbamic acid. 3) The esterification involves the 5'-hydroxy group of thymidine.

    \vspace{4pt}
    
    \textbf{Step 3: Construct the Molecular Structure}

    \vspace{4pt}
    
    The structure involves a thymidine molecule with a specific ester and carbamate group attached to the 5'-hydroxy group. The structure will have a complex arrangement of carbon, oxygen, and nitrogen atoms with multiple bonds and branching points.

    \vspace{4pt}
    
    \textbf{Step 4: Generate the SELFIES Representation}

    \vspace{4pt}
    
    The SELFIES representation is a compact and efficient way to encode molecular structures. It involves a series of tokens that represent the structure's components and bonds.

    \vspace{4pt}
    
    \textbf{Final Answer}: After constructing the molecular structure and generating the SELFIES representation, the final answer is: 
    
    \quad $\texttt{[C][C@H1][C@@H1][Branch2][\#Branch1][Ring2][C@H1][Branch2][=Branch1][\#C][C@H1]}$
    
    \quad $\texttt{[Branch2][=Branch1][\#Branch2][C@@H1][Branch1][Ring2][O][Ring1][=Branch1][O][C]}$
    
    \quad $\texttt{[C@@H1][C@H1][Branch2][Branch1][N][C@@H1][Branch2][Branch1][\#Branch1][C@H1]}$
    
    \quad $\texttt{[Branch2][Branch1][C][C@@H1][Branch1][Ring2][O][Ring1][=Branch1][O][C][=C]}$
    
    \quad $\texttt{[Branch2][Ring1][Ring2][O][C][=C][C][=Branch1][=N][=C][C][=Branch1][Branch2]}$
    
    \quad $\texttt{[=C][Ring1][=Branch1][O][C][O][O][O][O][O][O][O][O]}$
    
    \\\midrule

    \textbf{Innovator-Reason Answer} & 
    $\texttt{[C][C@H1][C@@H1][Branch2][\#Branch1][Ring2][C@H1][Branch2][=Branch1][\#C][C@H1]}$
    
    $\texttt{[Branch2][=Branch1][\#Branch2][C@@H1][Branch1][Ring2][O][Ring1][=Branch1][O][C]}$
    
    $\texttt{[C@@H1][C@H1][Branch2][Branch1][N][C@@H1][Branch2][Branch1][\#Branch1][C@H1]}$
    
    $\texttt{[Branch2][Branch1][C][C@@H1][Branch1][Ring2][O][Ring1][=Branch1][O][C][=C]}$
    
    $\texttt{[Branch2][Ring1][Ring2][O][C][=C][C][=Branch1][=N][=C][C][=Branch1][Branch2]}$
    
    $\texttt{[=C][Ring1][=Branch1][O][C][O][O][O][O][O][O][O][O]}$\\
    
    \bottomrule
  \end{tabular}
\end{table}

\end{document}